\documentstyle[colap,eepic,lingmacros,tree]{article}
\let\citeA=\newcite
\newcommand{\summary}[1]{\begin{abstract}#1\end{abstract}}
\newcommand{\subject}[1]{}\newcommand{\wordcount}[1]{}
\newcommand{\makeidpage}{}
\newcommand{\pref}[1]{(\ref{#1})}
\newcommand{\sect}[1]{Section~\ref{#1}}
\newcommand{\fig}[1]{Figure~\ref{#1}}
\newbox\bx

\long\def\enum[#1]#2{\enumsentence{\label{#1}#2}}
\long\def\eenum[#1]#2{%
\refstepcounter{enums}\label{#1}\eenumsentence[\theenums]{#2}}
\newcommand{\dnode}{\Node{$\vdots$}}

\newcommand{\expct}{{\cal E}}

\begin{document}
\title{Issues in Communication Game%
\vbox to0pt{\vss\rlap{\kern7em\raise3cm\llap{\normalsize\sl
Proceedings of COLING\,'96, pp.\,531-536.}}}}
\author{HASIDA K\^oiti\\Electrotechnical Laboratory\\
1-1-4 Umezono, Tukuba, Ibaraki 305, Japan.\\\sl hasida@etl.go.jp}
\maketitle
\summary{
As interaction between autonomous agents,
communication can be analyzed in game-theoretic terms.
{\em Meaning game} is proposed to formalize
the core of intended communication
in which the sender sends a message and the receiver
attempts to infer its meaning intended by the sender.
Basic issues involved in the game of natural language communication
are discussed, such as salience, grammaticality, common sense,
and common belief, together with some demonstration of the feasibility
of game-theoretic account of language.
}
\subject{discourse and pragmatics}
\wordcount{4000}
\makeidpage

\section{Introduction}\label{intro}

Communication is a game (interaction among autonomous agents) by definition.
So it can be analyzed in game-theoretic terms.\footnote{
See \citeA{osborne&rubinstein94}, among others,
for general reference on game theory.}
In this paper we study a fundamental aspect of linguistic communication
from the point of view of game theory,
and enumerate some basic issues involved in the communication
games of natural language.

Let $I$ be the proposition that the sender $S$ intends
to communicate a semantic content $c$ to the receiver $R$.
Then $I$ entails that $S$ intends that $R$ should both
recognize $c$ and believe $I$.
This is the core of {\em nonnatural meaning} \cite{grice57,grice69}.
Grice's original notion of nonnatural meaning further entails
($S$'s intention of) $R$'s believing (when $c$ is a proposition or a reference)
or obeying (when it is an order or a request) $c$,
but we disregard this aspect and concentrate on this core.

This restricted sense of nonnatural meaning
implies that communication is inherently collaborative,
because both $S$ and $R$ want that $R$ should recognize $c$ and $I$.
$S$ of course wants it, and so does $R$ because it is beneficial in general
to know what $S$ intends to make $R$ believe or obey.
$S$ might be lying or trying to mislead $R$,
but even in such a case $S$ is still intending to communicate a content $c$
by way of making $R$ recognize this intention.
Even if $R$ doubts $S$'s honesty, $R$ will try to know what $c$ is,
because knowing what $c$ is would help $R$ infer
what the hidden intent of $S$ may be, among others.
For instance, when $S$ tells $R$ that it is raining,
$R$ will learn that $S$ wants to make $R$ believe that it is raining.
$R$ would do so even if $R$ knew that it is not raining.
Even if $R$ were unsincere and misunderstood $S$'s message
on purpose,\footnote{
If $R$ is sincere and unintentionally misunderstands,
that is just a failure of sharing the same context with $S$.}
the nonnatural meaning is still properly conveyed,
because otherwise the intended misunderstanding would be impossible.

The present study concerns this aspect of communication,
the nonnatural meaning in the restricted sense,
which is a core of intended communication.
Lies, ironies, indirect speech acts, and so forth
\cite{perrault90,perrault&allen80} all share this core.
Our understanding about it will hence help us
understand basic workings of natural communication systems.
As an example, centering theory \cite{grosz&joshi&weinstein95}
could be attributed to game-theoretic accounts,
as demonstrated later in this paper.

\section{Communication Games}

\begin{figure}[htbp]
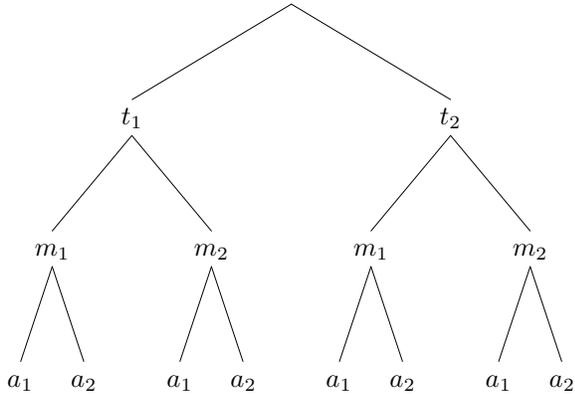

\begin{center}\tree{\unitlength=.14ex{
\Ln3{$t_1$}{
	\ln9{$m_1$}{
		\ln4{$a_1$}}
		\rn4{$a_2$}}
	\rn9{$m_2$}{
		\ln4{$a_1$}}
		\rn4{$a_2$}}
\Rn3{$t_2$}{
	\ln9{$m_1$}{
		\ln4{$a_1$}}
		\rn4{$a_2$}}
	\rn9{$m_2$}{
		\ln4{$a_1$}}
		\rn4{$a_2$}}
\end{center}
\caption{A signaling game.}
\label{signaling}
\end{figure}

Communication has been discussed in the game-theory literature.
A {\em signaling game} consists of
sender $S$'s sending a {\em message} (or a {\em signal}) to receiver $R$
and $R$'s doing some {\em action} in response to that message.
Here $S$ knows something that $R$ did not know before receiving the message.
This is formulated by assuming
that $S$ belongs to some {\em type},
which $S$ knows but $R$ does not know at first.
Let $T$ be the set of the types,
$P$ be the probability distribution over $T$.
Let $M$ be the set of the messages and
$A$ be the set of $R$'s possible actions.
Finally, let $U_X$ be the {\em utility function} for player $X$.
$U_S(t,m,a)$ and $U_R(t,m,a)$ are real numbers for
$t\in T$, $m\in M$ and $a\in A$.
A signaling game with
$T=\{t_1,t_2\}$, $M=\{m_1,m_2\}$ and $A=\{a_1,a_2\}$
is illustrated by a {\em game tree} as shown in \fig{signaling}.
Here the game proceeds downwards.
The top branch is the nature's initial choice of $S$'s type according to $P$,
the middle layer is $S$'s decision on which message to send,
and finally the bottom layer is $R$'s choice of her action.
When $R$ has just received $m_i$ ($i=1,2$), she does not know
whether the game has been played through $t_1$ or $t_2$.

Let $\sigma_S$ and $\sigma_R$ be $S$'s and $R$'s {\em strategies},\footnote{
Or {\em mixed strategies}, which are probability
distributions over the {\em simple strategies} (actions).} respectively.
That is, $\sigma_S(m|t)$ is the conditional probability of
$S$'s sending message $m$ provided that she is of type $t$,
and $\sigma_R(a|m)$ the conditional probability of $R$'s
doing action $a$ provided that she has received $m$.
The combination $\langle\sigma_S,\sigma_R\rangle$ of strategies is
an {\em equilibrium}\footnote{Or {\em complete Bayesian equilibrium},
in a more precise, technical term.} of a signaling game when
$\sigma_S$ and $\sigma_R$ are the optimal responses to each other;
that is, when $\sigma_X$ maximizes $X$'s expected utility
\[
\sum_{t,m,a}P(t)\,\sigma_S(m|t)\,\sigma_R(a|m)\,U_X(t,m,a)
\]
given $\sigma_Y$, for both $X=S\wedge Y=R$ and $X=R\wedge Y=S$.

In typical applications of signaling game,
$T$, $M$ and $A$ are not discrete sets as in the above example
but connected subsets of real numbers,
and $S$'s preference for $R$'s action is the same irrespective of her type.
In this setting, $S$ should send a costly message to get a large payoff.
For instance, in job market signaling \cite{spence73},
a worker $S$ signals her competence (type) to a potential employer $R$
with the level of her education as the message,
and $R$ decides the amount of salary to offer to $S$.
A competent worker will have high education and
the employer will offer her a high salary.
In mate selection \cite{zahavi75}, a deer $S$ indicates its strength by
the size of its antlers to potential mates $R$.
A strong deer will grow extra large antlers to demonstrate
its extra survival competence with this handicap.

{\em Cheap-talk game} is another sort of communication game.
It is a special case of signaling game
where $U_S$ and $U_R$ do not depend on the message; that is,
composing/sending and receiving/interpreting message are free of cost.
In a cheap-talk game, $S$'s preference for $R$'s action
must depend on her type for non-trivial communication to obtain,
because otherwise $S$'s message would give
no information to $R$ about her type.

\section{Meaning Game}\label{MG}

Now we want to formulate the notion of {\em meaning game}
to capture nonnatural meaning in the restricted sense
discussed in \sect{intro}.
Let $C$ be the set of semantic contents
and $P$ the probability distribution over the linguistic reference
to the semantic contents.
That is, $P(c)$ is the probability that
$S$ intends to communicate semantic content $c$ to $R$.
As before, $M$ is the set of the messages.
A meaning game addresses a {\em turn} of communication
$\langle c_S,m,c_R\rangle$, which stands for a course of events where
$S$, intending to communicate a semantic content $c_S$,
sends a message $m$ to $R$ and $R$ interprets $m$ as meaning $c_R$.
$c_S=c_R$ is a necessary condition 
for this turn of communication to be successful.
It seems reasonable to assume that the success
of communication is the only source of positive utility for any player.

So a meaning game might be a sort of signaling game
in which $S$'s type stands for
her intending to communicate some semantic content,
and $R$'s action is to infer some semantic content.
That is, both $T$ and $A$ could be simply regarded as $C$.
Strategies $\sigma_S$ and $\sigma_R$ are defined accordingly.

In a simple formulation,
the utility function $U_X$ of player $X$ would thus be a real-valued
function from $C\times M\times C$ (the set of turns).
It would be sensible to assume that
$U_X(c_S,m,c_R)>0$ holds only if $c_S=c_R$.
$U_X$ reflects the grammar of the language (which might be private
to $S$ or $R$ to various degrees).
The grammar evaluates the (computational, among others) cost
of using content-message pairs.
The more costly are $\langle c_S,m\rangle$
and $\langle m,c_R\rangle$, the smaller is $U_X(c_S,m,c_R)$.
The notion of equilibria in a meaning game is naturally derived from
that in a signaling game.

If the players want
something like {\em common belief},\footnote{People have common
belief of proposition $p$ when they all believe $p$, they all believe
that they all believe $p$, they all believe that
they all believe that they all believe $p$, and so on, ad infinitum.}
however, meaning games are not signaling games.
This is because $c_S=c_R$ is not a sufficient condition for
the success of communication in that case.
$U_X$ should then depend on not just $c_S$, $m$, and $c_R$,
but also the players' nested beliefs about each other.
We will come back to this issue in \sect{MB}.

Note also that the typical instances of meaning game in natural
language communication is not like the typical applications
of signaling game such as mentioned before,
even if meaning games are special sort of signaling games.
That is, meaning games in natural language would normally
involve discrete sets of semantic contents and messages.

Natural-language meaning games are not cheap-talk games, either,
because we must take into consideration the costs of content-message pairs.
It is not just the success of communication
but also various other factors that account for the players' utility.
$S$ and $R$ hence do not just want to
maximize the probability of successful communication.

\begin{figure}[htbp]
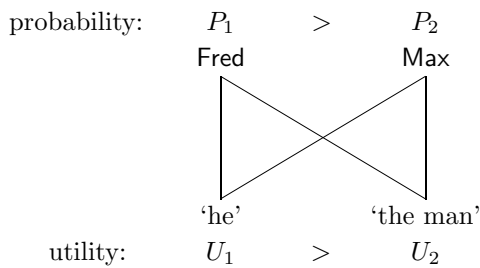

\begin{center}
\tree{\unitlength=.18ex\posx=100\unitlength{
\node{\llap{probability:\hspace{2.4em}}$P_1$}
\node{\sf Fred}{
	\lf{`he'}\node{\llap{utility:\hspace{3em}}$U_1$}}
	\Rn3{`the man'}\node{\llap{$>$\hspace{3em}}$U_2$}}
\advance\posx by100\unitlength
\node{\llap{$>$\hspace{3em}}$P_2$}
\node{\sf Max}{
	\Larc3}
	\varc}
\end{center}
\caption{A meaning game about references of NPs.}
\label{NPgame}
\end{figure}

To illustrate a meaning game and to demonstrate that meaning games
are not cheap-talk games, let us consider the following discourse.
\enum[he&man]{$u_1$: Fred scolded Max.\\
$u_2$: He was angry with the man.}
The preferred interpretation of `he' and `the man' in $u_2$
are {\sf Fred} and {\sf Max}, respectively, rather than the contrary.
This preference is accounted for
by the meaning game as shown in \fig{NPgame}.
In this game, {\sf Fred} and {\sf Max} are semantic contents,
and `he' and `the man' are messages.\footnote{
Perhaps there are other semantic contents and messages.}
We have omitted the nature's selection among the semantic contents.
Also, the nodes with the same label are collapsed to one.
$S$'s choice goes downward and $R$'s choice upward,
without their initially knowing the other's choice.
The complete bipartite connection between the contents and the messages
means that either message can mean either content grammatically
(without too much cost).

$P_1$ and $P_2$ are the prior probabilities
of references to {\sf Fred} and {\sf Max} in $u_2$, respectively.
Since {\sf Fred} was referred to by the subject and {\sf Max}
by the object in $u_1$,
{\sf Fred} is considered more salient than {\sf Max} in $u_2$.
This is captured by assuming $P_1>P_2$.
$U_1$ and $U_2$ are the utility (negative cost) of using
`he' and `the man,' respectively.\footnote{
For the sake of simplicity, here we assume that $U_S$ and $U_R$ are equal.
See \sect{MB} for discussion.}
Utilities are basically assigned to content-message pairs,
but sometimes it is possible to consider costs of messages
irrespective of their contents.
We assume $U_1>U_2$ to the effect that `he' is less complex than
`the man' both phonologically and semantically;
`he' is not only shorter than `the man' but also,
more importantly, less meaningful in the sense that
it lacks the connotation of being adult which `the man' has.

\begin{figure}[htbp]
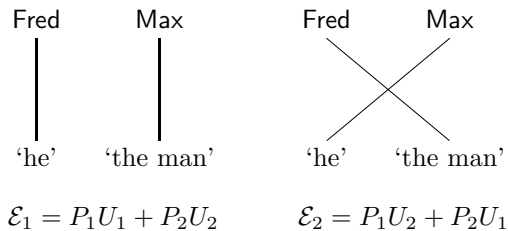

\begin{center}\tabcolsep=1.5em
\begin{tabular}{cc}
\tree{\unitlength=.15ex{
\node{\sf Fred}\lf{`he'}}
\advance\posx by72\unitlength
\node{\sf Max}\lf{`the man'}}
&
\tree{\unitlength=.15ex{
\node{\sf Fred}\Rn1{`the man'}}
\advance\posx by72\unitlength
\node{\sf Max}\Ln1{`he'}}\\
\\
$\expct_1=P_1U_1+P_2U_2$ & $\expct_2=P_1U_2+P_2U_1$
\end{tabular}
\end{center}
\setbox\bx\hbox{\fig{NPgame}}
\caption{Two equilibria of the meaning game in \usebox\bx.}\label{pcmue}
\end{figure}

There are exactly two equilibria entailing 100\%\ success of
communication, as depicted in \fig{pcmue} with their
expected utilities $\expct_1$ and $\expct_2$
apart from the utility of success of communication.\footnote{
Common belief about the communicated content is always obtained in both cases.
So the current discussion does not depend on whether the success
of communication is defined by $c_S=c_R$ or common belief.}
$P_1>P_2$ and $U_1>U_2$ imply $\expct_1-\expct_2=(P_1-P_2)(U_1-U_2)>0$.
So the equilibrium in the left-hand side
is preferable for both $S$ and $R$, or {\em Pareto superior}.
This explains the preference in \pref{he&man}.
It is straightforward to generalize this result for cases with more than
two contents and messages:
A more salient content should be referred to by a lighter message
when the combinations between the contents and the messages are complete.
A general conjecture we might draw from this discussion is the following.
\enum[pareto]{Natural-language meaning games are played at their
Pareto-optimal equilibria.}
An equilibrium is {\em Pareto optimal} iff
no other equilibrium is Pareto superior to it.

Note that we have derived an essence of centering theory
\cite{joshi&weinstein81,kameyama86,walker&iida&cote94,grosz&joshi&weinstein95}.
Centering theory is to explain anaphora in natural language.
It considers list $Cf(u_i)$ of {\em forward-looking centers},
which are the semantic entities {\em realized}\footnote{
A linguistic expression {\em realizes} a semantic content
when the former directly refers to the latter or the situation described
by the former involves the latter.
For instance, after an utterance of `a house,'
`the door' realizes the house referred to by `a house.'} in $u_i$,
where $u_i$ is the $i$-th utterance.
The forward-looking centers of utterance $u$ are ranked in $Cf(u)$
according to their saliences.
In English, this ranking is determined by grammatical functions
of the expressions in the utterance, as below.
\begin{quote}
subject $>$ direct object $>$ indirect object
$>$ other complements $>$ adjuncts
\end{quote}
The highest-ranked element of $Cf(u)$ is called the {\em preferred center}
of $U$ and written $Cp(u)$.
{\em Backward-looking center} $Cb(u_i)$ of utterance $u_i$
is the highest-ranked element of $Cf(u_{i-1})$ that is realized in $u_i$.
$Cb(u)$ is the entity which the discourse
is most centrally concerned with at $u$.

Centering theory stipulates the following rule.
\enum[RuleI]{
If an element of $Cf(u_{i-1})$ is realized by a pronoun in $u_i$,
then so is $Cb(u_i)$.}
In \pref{he&man}, $Cb(u_2)=$ {\sf Fred}
because $Cf(u_1)=$ [{\sf Fred}, {\sf Max}],
if either `he' or `the man' refers to {\sf Fred}.
Then rule \pref{RuleI} predicts that {\sf Fred} cannot be realized by `the man'
if {\sf Max} is realized by `he'
--- the same prediction that we derived above.
Moreover, \pref{RuleI} itself is a special instance
of our above observation that 
a more salient content should be referred to by a lighter message,
provided that the backward-looking center is particularly salient.

\pref{RuleI} is common in all the version of centering theory,
but of course there are further details of the theory,
which vary from one version to another.
To derive all of them (which are right)
in a unified manner requires further extensive study.

\section{Playing the Same Game}\label{MB}

We have so far assumed implicitly that $S$ and $R$ have common knowledge
about (the rule of) the game (that is, $P$, $U_S$ and $U_R$).
This assumption will be justified as a practical approximation
in typical applications of signaling games (and cheap-talk games).
For instance, there may well be a body of roughly
correct, stable common-sense knowledge
about the correlation between the competence
of workers and the degree of effort they make to have higher education,
about how much an employer will offer to
an employee with a certain competence, and so on.

However, common knowledge on the game might be harder to obtain
in natural-language meaning games, because the game
lacks such stability of the typical signaling games as mentioned above.
A natural-language meaning game is
almost equivalent to the context of discourse,
which changes dynamically as the discourse unfolds.

\begin{figure*}[htbp]
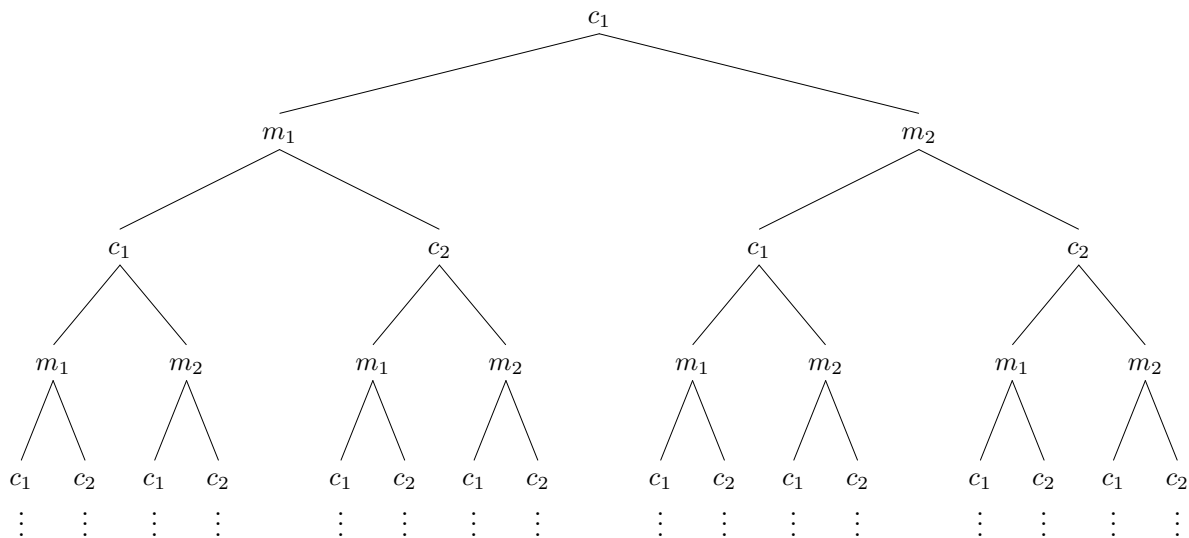

\def\ONE{{\Ln7{$m_1$}\TWO}\Rn7{$m_2$}\TWO}
\def\TWO{{\Ln4{$c_1$}\THREE}\Rn4{$c_2$}\THREE}
\def\THREE{{\ln9{$m_1$}\FOUR}\rn9{$m_2$}\FOUR}
\def\FOUR{{\ln5{$c_1$}\FIVE}\rn5{$c_2$}\FIVE}
\def\FIVE{\dnode}
\begin{center}\tree{\unitlength=.117ex
\node{$c_1$}\ONE}
\end{center}
\caption{Inference by $S$ to communicate semantic content $c_1$.}
\label{tgame}
\end{figure*}

In general, to figure out her own best strategy,
$S$ ($R$) attempts to infer $R$'s ($S$'s) strategy
by simulating $R$'s ($S$'s) inference.
If $S$ and $R$ do not have common knowledge about the game,
this inference will constitute an infinite tree.\footnote{
This is not a game tree but a tree of belief embedding.}
For instance, \fig{tgame}
depicts $S$'s inference when she wants to communicate $c_1$,
where the players have common knowledge
of $C=\{c_1,c_2\}$ and $M=\{m_1,m_2\}$ but not of their utility functions.
The nodes labeled by $c_i$ represent $S$ when she wants to communicate
$c_i$, and those labeled by $m_i$ represent $R$ when she wants
to interpret $m_i$, for $i=1,2$.
The inference by $R$ when interpreting message $m_i$
is a similar tree rooted by $m_i$.

Although it is impossible to actually have common knowledge
in general \cite{halpern&moses90},
there are several motivations for the players to pretend
to have common knowledge about the game.
First, they can avoid the computational complexity in dealing with
infinite trees such as above.
Second, common belief on the game is a simple means to obtain common belief
on the communicated content.
Third, the best payoff is obtained
when the players have common knowledge about the game,
if their utility functions are equal.
In fact, the utility functions are probably equal,
because language use as a whole is a {\em repeated game}.
That is, provided that communicating agents
play the role of $S$ and $R$ half of the time each,
they can maximize their expected utility by
setting their utility functions to the average of their selfish utilities.
Fortunately, this equalization is very stable,
as long as the success of communication is the only source
of positive utility for both the players.

In communication games, common knowledge on which message $S$ has sent
should help the players converge on common belief on the game.
That is, when the players have common knowledge that message $m$
was sent, they may be able to detect errors in their embedded beliefs.
In fact, an embedded belief turns out wrong
if it implies $\sigma_S(m|c)=0$ for every $c$ in the embedded context.
This common knowledge about $m$ may be even incorporated in the meaning game.
That is, it may affect the cost of retrieving or composing
various content-message pairs,
thus biasing the scope of the game towards
those content-message pairs closely associated with $m$.
Contents and messages very difficult to envisage given $m$
will be virtually excluded from the game.
Once the game is defined, however, 
both players must take into consideration
the entire maximal connected subgraph containing the content
she wants to convey or the message she wants to interpret.

\section{Composite Game}

Natural-language communication is a composite game in two senses.
First, as mentioned in the previous section,
it is considered a {\em repeated game}, which is a sequence of smaller games.
Second, each such smaller game is a compound game
consisting of temporally overlapping meaning games.
These facts introduce several complications into the communication game.

In a repeated game, one stage may affect the subsequent stage.
In natural-language communication,
a meaning game can influence the next meaning game.
For instance, if a semantic content $c$ is referred to by
a message with a low cost,
then the probability of reference to $c$ may increase
as a sort of accommodation,\footnote{
\citeA{lewis79} discusses several types of accommodation for
conversational score, of which the most relevant here is accommodation
for comparative salience:
$x$ becomes more salient than $y$ when something is said which presupposes
$x$ to be more salient than $y$.}
because a reference by a lightweight message presupposes
high prior probability of reference, as discussed in \sect{MG}.
For instance, a reference to {\sf Fred} by `he' will raise
the salience of {\sf Fred}.

Another type of contextual effect shows up the following discourse.
\enum[man&him]{
$u_1$: Fred scolded Max.\\
$u_2$: The man was angry with him.}
Here `the man' and `he' in $u_2$ are more readily interpreted
as {\sf Fred} and {\sf Max}, respectively, which violates \pref{RuleI}
and hence our game-theoretic account.

\begin{figure}[htbp]
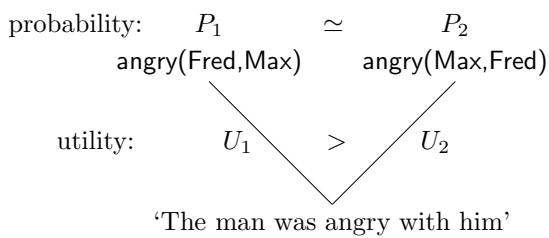

\begin{center}\tree{\unitlength=.18ex\posx=92\unitlength{
\node{\llap{probability:\hspace{2em}}$P_1$}
\node{\sf angry(Fred,Max)}{
	\Rn0{`The man was angry with him'}}}
\advance\posx by60\unitlength{
\advance\posy by-45\unitlength
\node{\llap{utility:\hspace{3em}}
\leavevmode\hbox to20ex{$U_1$\hss$>$\hss$U_2$}}}
\advance\posx by60\unitlength
\node{\llap{$\simeq$\hspace{8.5ex}}$P_2$}
\node{\sf angry(Max,Fred)}
	\Larc0}
\end{center}
\caption{A meaning game about propositions and sentences.}\label{psgame}
\end{figure}

This preference is accounted for by the preference for parallelism
concerning the combination of semantic content and grammatical function:
In both $u_1$ and $u_2$ {\sf Fred}
is realized by the subject NP and
{\sf Max} is realized by the object NP.
This is the same sort of preference that is addressed by
property-sharing constraint \cite{kameyama86}.
This effect is attributed to the utility assignment
as shown in \fig{psgame}.
That is, the utility $U_1$ of associating the proposition
{\sf angry(Fred,Max)} (that Fred is angry with Max)
with the sentence `The man was angry with him'
is greater than the utility $U_2$ of
associating {\sf angry(Max,Fred)}
(the proposition that Max is angry with Fred) with the same sentence.
This game might involve other possible associations
such as that between {\sf angry(Max,Fred)} and `The man made him angry,'
but as mentioned at the end of \sect{MB} contents
and messages other than included in \fig{psgame} probably accompany
great costs and hence may be neglected.

In general, several meaning games are played possibly in parallel during
linguistic communication using a compound expression.
A turn of communication with an utterance of `the man was angry with him'
consists of the sentence-level game mentioned above,
the two noun phrase-level games --- one concerning the subject NP
(shown in \fig{NPgame}) and the other the object NP of `with' --- and so on.
A strategy of each player
in such a compound game associated with a compound expression
is a combination of her strategies for all such constituent games.
Each player attempts to maximize the expected utility
over the entire compound game, rather than for each constituent game.

Different constituent games often interact.
For instance, if the speaker chooses to say `the man' for the subject NP,
then the whole sentence cannot be `he was angry with the man.'
So a global solution, which maximizes the utility from the entire game,
may maximize the utility from some constituent games but not from others.
In the above example, the global solution, which involves
saying `the man was angry with him' and interpreting it as
{\sf angry(Fred,Max)}, maximizes the utility from the sentence-level game
but not from the NP-level games.
Incidentally, the players will gain greater utility
if they use the combination of {\sf angry(Fred,Max)}
and `he was angry with the man,'
which is consistent with the optimal equilibrium of the NP-games.
When `the man was angry with him' is used despite the smaller default
utility associated with it,
{\sf Max} will probably be assigned a greater salience than otherwise,
which is again a sort of accommodation.

Extralinguistic context enters sentence-level games
and plays an important role in language use.
For example, if it is known that
Max never gets angry and that Fred is short-tempered,
then both in \pref{he&man} and \pref{man&him}
the second utterance will preferably
be interpreted as meaning {\sf angry(Fred,Max)}.

\section{Conclusion}

{\em Meaning game} captures nonnatural meaning in the restricted sense
which obtains in basically all the cases of natural-language communication.
The factors which define a meaning game include
grammatical function, reference by lightweight message,
extralinguistic information (these affect $P$),
grammar, cost of recalling (these affect the utility), and so on.
To have a more complete game-theoretic account of natural language,
we need a quantitative characterization of how those factors
contribute to the game.

We have almost restricted ourselves to references of nouns phrases,
but the seemingly promising targets of game-theoretic account
of natural language apparently include
binding theory, conversational implicature \cite{pparikh92}, and so on.
Since our account is very general in nature, however,
it should apply to language as a whole.
For instance, the notion of grammaticality may well be attributed to
the computational difficulty in convergence to
a common game-theoretic equilibrium.
Also, abductive inference involved in language understanding
\cite{hobbs&93} (hence in language production, too,
from the game-theoretic, reciprocal viewpoint)
is closely related with our theory.
That is, the particular usefulness of
abduction in natural language communication could be
ascribed to the fact that language use in essence is
a collaborative interaction such as discussed so far.

\bibliography{roman}
\bibliographystyle{acl}
\end{document}